\documentclass[letterpaper, 10 pt, journal, twoside]{IEEEtran}
\usepackage{amsmath,amsfonts}
\usepackage{array}
\usepackage[caption=false,font=normalsize,labelfont=sf,textfont=sf]{subfig}
\usepackage{textcomp}
\usepackage{stfloats}
\usepackage{url}
\usepackage{verbatim}
\usepackage{graphicx}
\usepackage{cite}
\usepackage[dvipsnames]{xcolor}
\usepackage[ruled]{algorithm2e}
\usepackage{algpseudocode}
\usepackage{graphics}
\usepackage{amsmath}

\usepackage{amssymb}
\usepackage{epsfig} 
\usepackage{booktabs}
\usepackage{float}
\usepackage{url}
\usepackage{gensymb}
\usepackage{hyperref}
\hypersetup{
    colorlinks=true,
    urlcolor=red,
    }
\usepackage{tikz}
\usepackage{color}
\definecolor{darkgreen}{RGB}{0,127,0}
\definecolor{darkblue}{RGB}{0,0,175}

\newcommand{\remindtext}[2]{{{#2}}}
\newcommand{\addedtext}[2]{{{#2}}}
\newcommand{\modifiedtext}[2]{{{#2}}}
\newcommand{\deletedtext}[2]{{}}

\newcount\Comments 
\usepackage{array,multirow,graphicx}
\usepackage{color}
\usepackage{wrapfig}
\definecolor{darkgreen}{rgb}{0,0.5,0}
\definecolor{purple}{rgb}{1,0,1}
\newcommand{\kibitz}[2]{\ifnum\Comments=1\textcolor{#1}{#2}\fi}

\newenvironment{mylist}
{\begin{list}{$\bullet$}
    {\leftmargin4mm \itemsep0pt \itemindent-2mm \topsep0mm \parsep0.5mm \labelsep2mm}
}
{
    \end{list}
}

\begin{document}

\title{ChatEMG: Synthetic Data Generation to Control a Robotic Hand Orthosis for Stroke}

\author{
Jingxi Xu$^{1, *}$, Runsheng Wang$^{2, *}$, Siqi Shang$^{1, *}$, Ava Chen$^2$, Lauren Winterbottom$^3$, To-Liang Hsu$^1$, \\Wenxi Chen$^2$, Khondoker Ahmed$^1$, Pedro Leandro La Rotta$^2$, Xinyue Zhu$^1$, \\Dawn M. Nilsen$^{3,4}$, Joel Stein$^{3,4}$, and Matei Ciocarlie$^{2,4}$
\\ \href{https://jxu.ai/chatemg}{https://jxu.ai/chatemg}%
\thanks{Manuscript received: June, 15, 2024; Revised September, 22, 2024; Accepted November, 10, 2024.}%
\thanks{This paper was recommended for publication by Editor Pietro Valdastri upon evaluation of the Associate Editor and Reviewers' comments.
This work was supported in part by the National Institutes of Health (R01NS115652, F31HD111301) and the CU Data Science Institute.}%
\thanks{$^*$These authors have contributed equally to this work.}%
\thanks{$^{1}$Department of Computer Science, Columbia University, New York, NY 10027, USA.
        {\tt\small jxu@cs.columbia.edu}, {\tt\small \{siqi.shang, th2881, kfa2122, xz3013\}@columbia.edu}}%
\thanks{$^{2}$Department of Mechanical Engineering,
        Columbia University, New York, NY 10027, USA.
        {\tt\small \{rw2967, ava.chen, wc2746, pll2127, matei.ciocarlie\}@columbia.edu}}%
\thanks{$^{3}$Department of Rehabilitation and Regenerative Medicine, Columbia University Irving Medical Center, New York, NY 10032, USA.
{\tt\small \{lbw2136, dmn12, js1165\}@cumc.columbia.edu}}%
\thanks{$^4$ Co-Principal Investigators}%
\thanks{Digital Object Identifier (DOI): see top of this page.}
}

\markboth{IEEE Robotics and Automation Letters. Preprint Version. Accepted November, 2024}%
{Xu \MakeLowercase{\textit{et al.}}: ChatEMG: Synthetic Data Generation to Control a Robotic Hand Orthosis for Stroke}


\maketitle

\begin{abstract}
    
    \modifiedtext{16-2}{Intent inferral on a hand orthosis for stroke patients is challenging due to the difficulty of data collection.} Additionally, EMG signals exhibit significant variations across different conditions, sessions, and subjects, making it hard for classifiers to generalize. Traditional approaches require a large labeled dataset from the new condition, session, or subject to train intent classifiers; however, this data collection process is burdensome and time-consuming. In this paper, we propose ChatEMG, an autoregressive generative model that can generate synthetic EMG signals conditioned on prompts (i.e., a given sequence of EMG signals). ChatEMG enables us to collect only a small dataset from the new condition, session, or subject and expand it with synthetic samples conditioned on prompts from this new context. ChatEMG leverages a vast repository of previous data via generative training while still remaining context-specific via prompting. Our experiments show that these synthetic samples are classifier-agnostic and can improve intent inferral accuracy for different types of classifiers. We demonstrate that our complete approach can be integrated into a single patient session, including the use of the classifier for functional orthosis-assisted tasks. To the best of our knowledge, this is the first time an intent classifier trained partially on synthetic data has been deployed for functional control of an orthosis by a stroke survivor.
\end{abstract}
\begin{IEEEkeywords}
Generative AI, Synthetic Data, Rehabilitation Robotics, Prosthetics and Exoskeletons, Wearable Robotics.
\end{IEEEkeywords}

\section{Introduction}

\modifiedtext{6-1a}{Wearable rehabilitation robots continuously interact with human patients and must constantly make decisions on when and how to provide motor assistance. Machine learning methods are being applied increasingly to mediate this stream of information.} However, many current learning methods, which have revolutionized domains such as vision or language, rely on the availability of large training datasets. \modifiedtext{6-1b}{Compared to these domains, applications for learning on wearable robots are faced with a tremendous scarcity of both raw data and reliable ground truth labels~\cite{meeker2017emg,xu2022adaptive,la2024meta}.}

\begin{figure}
    \centering
    \includegraphics[width=\linewidth]{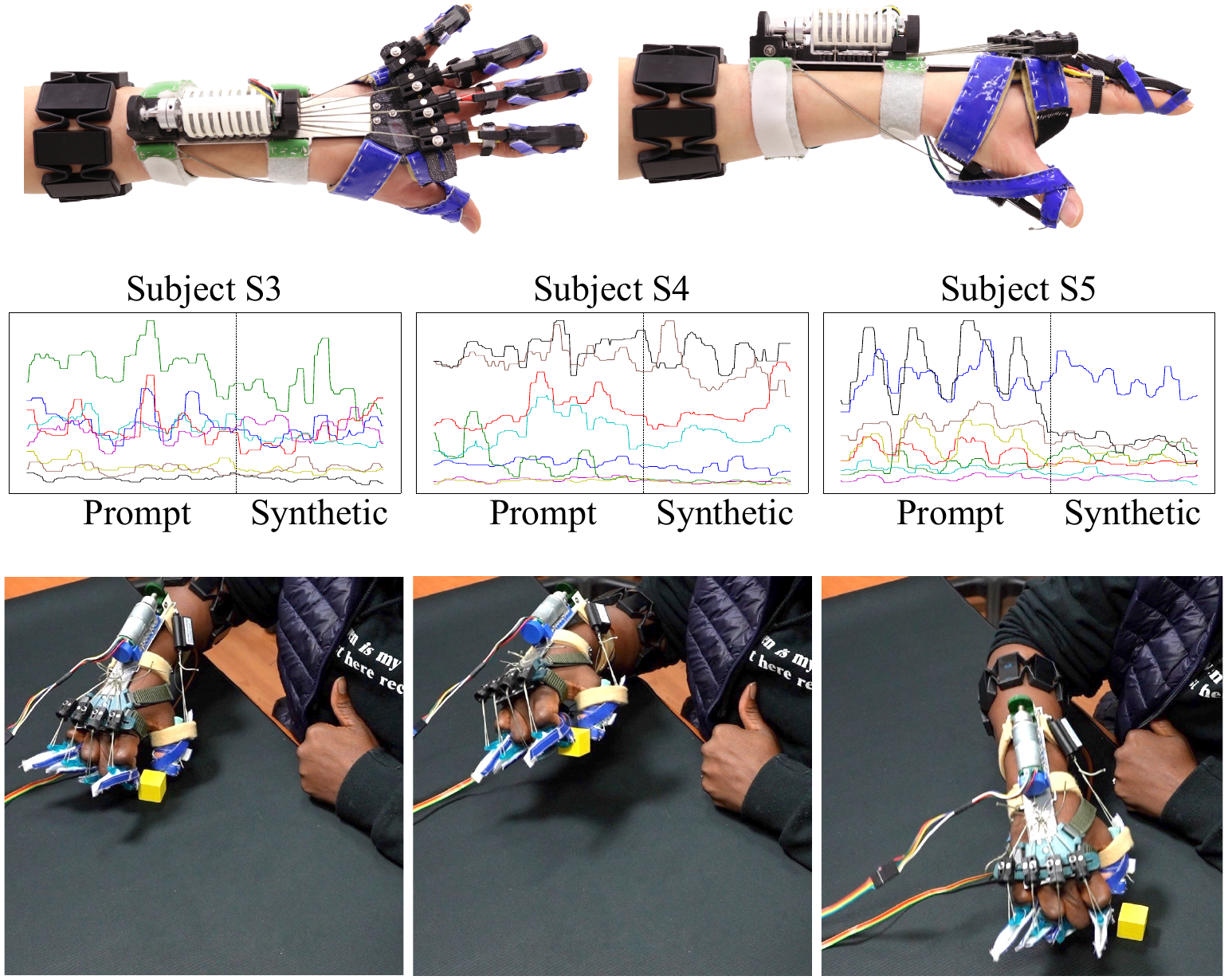}
    \vspace{-0.25in}
    \caption{Approach overview. Our hand orthosis (top row) collects EMG data from a forearm armband and uses this data to infer the patient's intent. ChatEMG models trained on a large corpus of offline data can generate synthetic data (middle row) for a new patient conditioned on a prompt from a small dataset of the new patient, and specific to an intended arm movement. The synthetic and real data are then used to jointly train an intent classifier, which, in the course of the same session, enables functional pick-and-place tasks (bottom row) with the orthosis.}
    \label{fig:teaser}
    \vspace{-0.25in}
\end{figure}

One such example is intent inferral, or the process by which a robotic orthosis or prosthesis collects a set of biosignals from the user, and uses them to infer the activity that the user intends to perform, so it can provide physical assistance at the right moment. In the case of a hand orthosis for stroke survivors developed in our lab~\cite{park2018multimodal,chen2022thumb} (Fig.~\ref{fig:teaser}), the device can use forearm electromyographic (EMG) data to predict when the user is trying to open the hand, and provide assistance to overcome the muscle spasticity.
Widely considered to be a key problem in assistive and rehabilitative robotics~\cite{beckerle2017}, an effective intent inferral mechanism can be an intuitive way to control a robotic device.

Like many applications in the assistive and rehabilitation devices domain, a fundamental challenge in intent inferral for stroke is the difficulty of collecting training data. The variation in EMG signals across conditions, sessions, and subjects makes the challenge even more pronounced. 
\modifiedtext{16-3}{
Firstly, EMG signal presents different patterns across subjects for the same intent due to variations in neuromuscular control impairments~\cite{tang2018surface,kyranou2018causes}.
In addition, defining a use \textit{session} as a single use of the device between donning and doffing, even for the same subject, the muscle tone and spasticity can vary across different sessions~\cite{la2024meta}. Furthermore, the signals are non-stationary and could change over time within a single session, depending on the use \textit{conditions}, such as the hand position and whether the motor is engaged and providing active grasping assistance~\cite{meeker2017emg,xu2022adaptive}.} Due to such variation, intent classifiers trained for a specific condition/session/subject, do not generalize well, and classical solutions often require tedious data collection on every new condition/session/subject, introducing a significant burden on the participants. 

In this work, we aim to reduce the burden of data collection from stroke subjects by generating synthetic data. We propose ChatEMG, an autoregressive generative model that understands the broad behavior of forearm EMG signals from a corpus of offline data across different stroke subjects and then can generate \textit{personalized} (i.e., condition-, session-, and subject-specific) synthetic samples conditioned on \textit{prompts} (i.e., a given sequence of EMG signals) sampled from a very limited dataset of a new condition, session, or subject. 


ChatEMG is Transformer-based, trained autoregressively, and temporal in its generative nature, meaning that each block of generated signals is conditioned on the previous blocks. The ability of ChatEMG to condition on a given sequence of EMG signals to generate synthetic signals of unlimited length is crucial to our application. Due to the significant variations of the EMG signals, the synthetic samples have to be personalized in order to be useful. As a result, ChatEMG leverages experience from previous data and produces synthetic samples conditioned on new data. In summary, our contributions are as follows:

\begin{mylist}
    \item We propose ChatEMG, an approach for producing synthetic EMG data via generative training on data also collected from stroke patients. Unlike previous models, ChatEMG can be conditioned on limited data from a new condition/session/subject in order to generate personalized synthetic sequences of arbitrary lengths. 
    \item We show that data generated by ChatEMG improves intent inferral performance for a broad range of intent classifiers. To the best of our knowledge, this is the first time that synthetic data has been shown to improve intent inferral performance when using real data from stroke patients.
    \item Our complete new patient protocol (collecting limited new data, using ChatEMG to generate personalized synthetic data, and then training an intent classifier), can be integrated into a single patient session. This increases the applicability of our method for functional tasks with real-world patients. To the best of our knowledge, this is the first time that an intent classifier trained partially on synthetic data has been deployed for functional orthosis control by a stroke patient.
\end{mylist}
\section{Related Work}

\paragraph{Intent Inferral with EMG Signals}
There are many previous works that attempt to infer the activity intents of disabled-bodied subjects using EMG signals with machine learning. These predicted intents are often sent to control wearable assistive and rehabilitative devices. 
A majority of works~\cite{meeker2017emg,castellini2009surface,powell2013training,lee2010subject} use supervised learning to control prostheses or orthoses, where the model is trained only on an initial dataset and then used during a longer session. However, it is not easy to collect labeled data on stroke subjects, and a new dataset needs to be collected for every new session after donning and doffing the device. Recent works~\cite{xu2022adaptive,zhai2017self,amsuss2013self,edwards2016application} 
have been exploring the semi-supervised learning paradigm to control wearable devices with EMG signals. This paradigm leverages unlabeled data to make the intent inferral algorithm more robust to change from the input signals within a single session. It still relies on some heuristics to ``label'' the unlabeled data, which can be inaccurate from time to time. 

\paragraph{Generative AI in Biomedical Research}


The lack of accurate and reliable data is not specific to intent inferral, but to the general machine learning research in the medical community. Similar to ours, there is some previous work that studies generative models for synthetic data generation. A majority of them~\cite{shin2018medical,hazra2020synsiggan,anicet2020parkinson,coelho2023novel,piacentino2021generating,festag2022generative,yang2023ts} use Generative Adversarial Network (GAN)~\cite{goodfellow2014generative} and its variants. 


GAN and its variants are not autoregressive and can only generate a fixed length of signals from random seeds. In comparison, ChatEMG can generate an unlimited sequence of EMG signals conditioned on EMG prompts, which are also of arbitrary length. This is extremely useful in applications where personalization is necessary. Stroke subjects exhibit different hand functionality 
and ChatEMG allows us to bias data generation via prompting. 

\modifiedtext{6-1c}{Most similar to ours, Bird et al.~\cite{bird2021synthetic} train a GPT-2 model to generate EMG signals for hand open/close classification for healthy subjects. However, their method also does not allow conditioning on prompts, and they only study random forest classifiers with healthy subjects. We specifically focus here on stroke patients whose intents are much more challenging to infer
and the signal variations across subjects make conditional data generation from prompts critical.}



\section{Overview}
\label{sec:overview}

\begin{figure*}
    \centering
    \includegraphics[width=\textwidth]{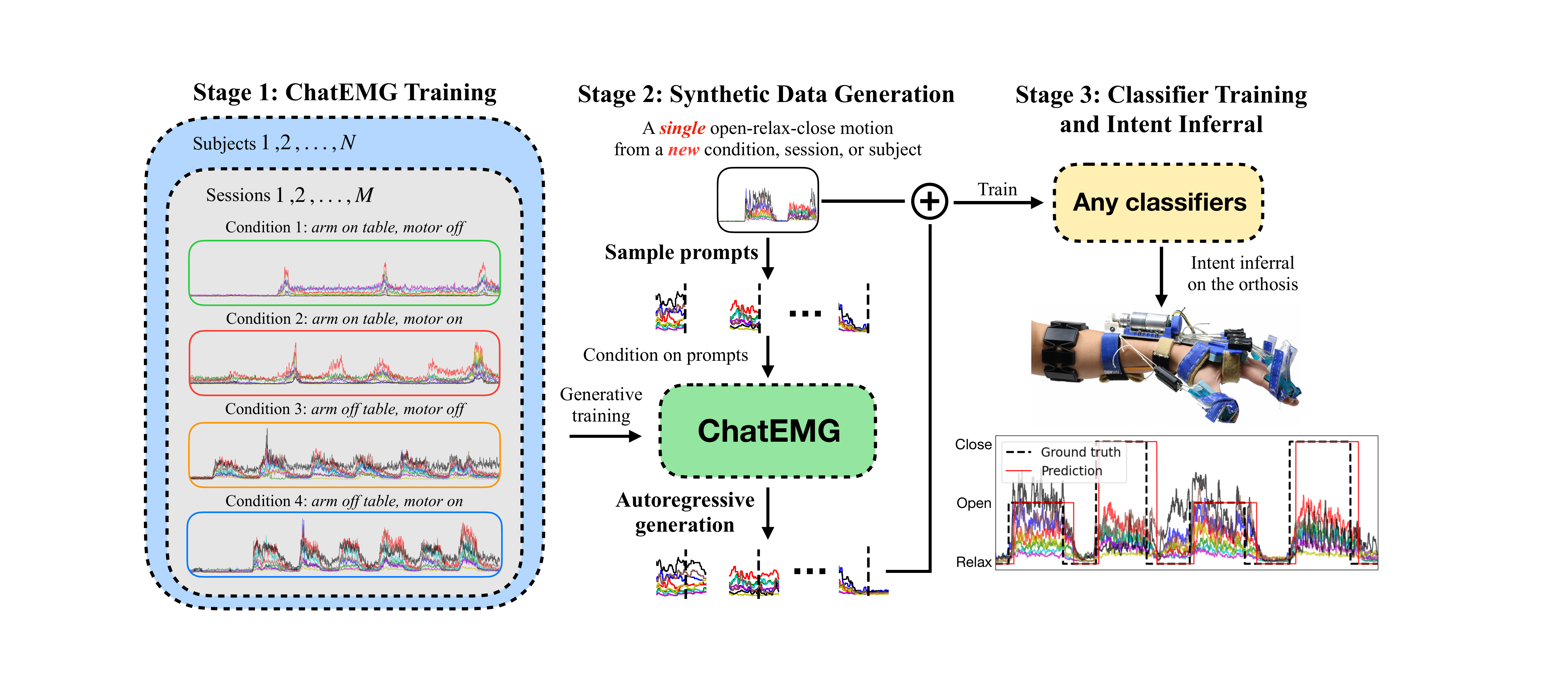}
    \vspace{-0.25in}
    \caption{\textbf{ChatEMG overview.} Stage 1: ChatEMG is trained on large offline data from different conditions, sessions, and subjects. We visualize the EMG recordings from different conditions of the same subject in a single session. \remindtext{16-3}{As shown here, there is a drastic variation in EMG signals for different conditions.} Stage 2: we only need a very limited labeled dataset from a new condition, session, and subject, and we use ChatEMG to expand this limited dataset with synthetic samples. These synthetic samples are conditioned on prompts from the new condition/session/subject. Stage 3: we train intent inferral classifiers using both the synthetic samples and the original limited dataset. Running the classifier, our orthosis can then provide active assistance for the stroke subjects in functional tasks.}
    \label{fig:pipeline}
    \vspace{-0.15in}
\end{figure*}

The ultimate goal of this project is to develop intent inferral classifiers that can predict stroke subjects' intent, so that our orthosis can provide meaningful functional assistance. Specifically, based on EMG data, we aim to predict which movement out of $\{open,close,relax\}$ the subject intends to perform with their hand. If the classifier predicts that the user intends to open, the device retracts the tendon, extending the fingers. If the user intends to close, the device extends the tendon, allowing the user to use their own grip strength to close their hand. If the predicted intent is to relax, the device maintains its previous state.

The most direct way to achieve this goal is to collect a set of training data (EMG signals) labeled with ground truth intent. This can be done by instructing the patient to attempt one of the three hand movements of interest, while simultaneously recording EMG data and labeling it with the prescribed intent. Once enough labeled data is collected for each of the three possible intent classes, we can train a classifier to distinguish between them. 

However, this traditional approach suffers from two key limitations: (1) The process of collecting labeled training data is burdensome and time-consuming for both the patient and the experimenter. It uses up precious session time and also fatigues the patients, leading to increased muscle tone and spasticity. (2) EMG signals exhibit significant variations between different conditions, sessions, and subjects. Training data collected in one condition/session/subject is unlikely to apply to a different one, leading to very poor generalization performance of the classifier. 

Our approach aims for a different paradigm. Its goal is to quickly adapt to a new condition, session or subject, using only a very small amount of newly collected, labeled training data. To achieve that, it relies on synthetic data from a generative model trained on a very large corpus of previously collected labeled data from a variety of conditions, sessions, and subjects. Concretely, our approach consists of three stages, illustrated in Fig.~\ref{fig:pipeline} and described below:

\subsubsection{ChatEMG Generative Training on Large Offline Data} In the first stage, a number of generative ChatEMG models are trained on a large corpus of offline data $\mathcal{D}_{\text{offline}}$ collected from different stroke subjects, which includes various conditions and sessions. One ChatEMG model is trained for each intent (open/close/relax). Once trained, each such model is able to generate synthetic data matching its respective intent. 

\subsubsection{Synthetic Data Generation Conditioned on Small Prompts} When a new condition/session/subject is started, we collect a very small labeled dataset $\mathcal{D}_{\text{new}}^{\text{orig}}$ in the new setting. We then use the ChatEMG models to extend this dataset through synthetic data generation. Concretely, for each possible intent, we use its respective model to generate additional synthetic data, referred to as $\mathcal{D}_{\text{new}}^{\text{synth}}$. 

Critically, $\mathcal{D}_\text{new}^\text{synth}$ is generated by models trained on $\mathcal{D}_{\text{offline}}$, but prompted with data sampled from $\mathcal{D}_\text{new}^\text{orig}$. The autoregressive nature of ChatEMG enables it to generate synthetic data of unlimited length conditioned on an existing piece of EMG sequence, which we call prompts. The ability to condition on prompts means that our synthetic data is based on knowledge mined from a large repository of previous data but applied in the context of the current condition, session, or subject. \textit{This is the essence of ChatEMG: it can leverage a vast repository of previous data via generative training while still remaining condition-, session- and subject-specific via prompting.}
    
\subsubsection{Classifier Training and Intent Inferral} Once the personalized synthetic data has been generated, we are ready to train an intent inferral classifier for the current situation. We train this classifier on both $\mathcal{D}_\text{new}^\text{orig}$ and $\mathcal{D}_\text{new}^\text{synth}$. We can then use this classifier for live intent inferral on our orthosis. 

It is worth noting that our approach is agnostic to the type of classifier used here. As we will show in the results section, this approach can be used with a variety of classifier architectures and generally improves their performance.
\section{ChatEMG Generative Models}

\begin{figure}[t]
    \centering
    \includegraphics[width=\linewidth]{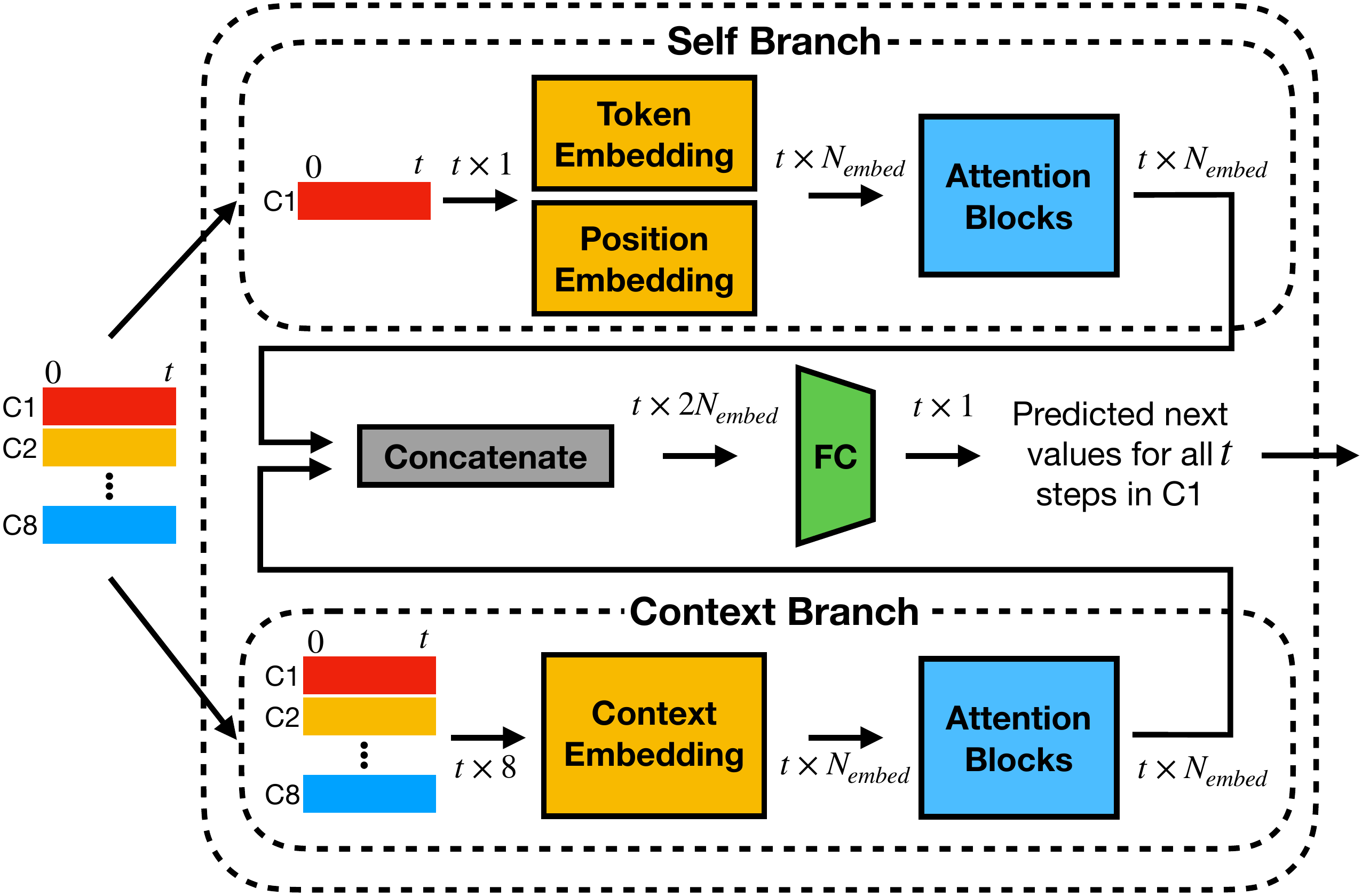}
    \vspace{-0.25in}
    \caption{\textbf{ChatEMG model architecture.} ChatEMG has two branches: the self branch that takes in the first channel (C1) and the context branch that takes in all 8-channel EMG signals. 
    }
    \label{fig:architecture}
    \vspace{-0.2in}
\end{figure}

The role of a ChatEMG model is to take in a sequence of EMG signals as input and predict the next EMG signal as output, where one signal consists of the 8-channel data from our EMG armband. \addedtext{16-4}{We use the Myo armband from the Thalmic Labs, which has 8 electrodes covering the forearm and collects signals at 100Hz.} Such a model can then be used autoregressively to generate synthetic data of arbitrary length, all conditioned on the given prompt. We note that this approach is similar in concept to language models such as ChatGPT~\cite{chatgpt}, capable of generating text in response to a text prompt. However, ChatEMG operates on the ``language'' of EMG data, hence its chosen moniker.

We train one ChatEMG model on data corresponding to each user intent. The goal is then for each of these models to generate synthetic data corresponding to the intent on which it has been trained. While each of these models is trained on different data, their architecture is the same.

\subsection{Architecture}

Each ChatEMG model is a Transformer-based decoder-only model with only self-attention mechanisms, similar to ChatGPT, as shown in Fig.~\ref{fig:architecture}. The input to this model consists of a time sequence of EMG signals. The attention mechanism of Transformers allows the input to be arbitrary length. Given an input sequence of length $t$, the output is a vector, also of length $t$, which contains the predicted next values for all $t$ steps in the first channel. ChatEMG only predicts the EMG value of the first channel such that the output value follows a discrete vocabulary of limited size, which we discuss later in this section. The $i$th element of this output vector is the predicted $(i+1)$th element of the first channel, with attention up to the $i$th signals in the input sequence. Details of the attention mechanism can be found in Vaswani et al.~\cite{vaswani2017attention}. 

\begin{figure}[t!]
    \centering
    \includegraphics[width=\linewidth]{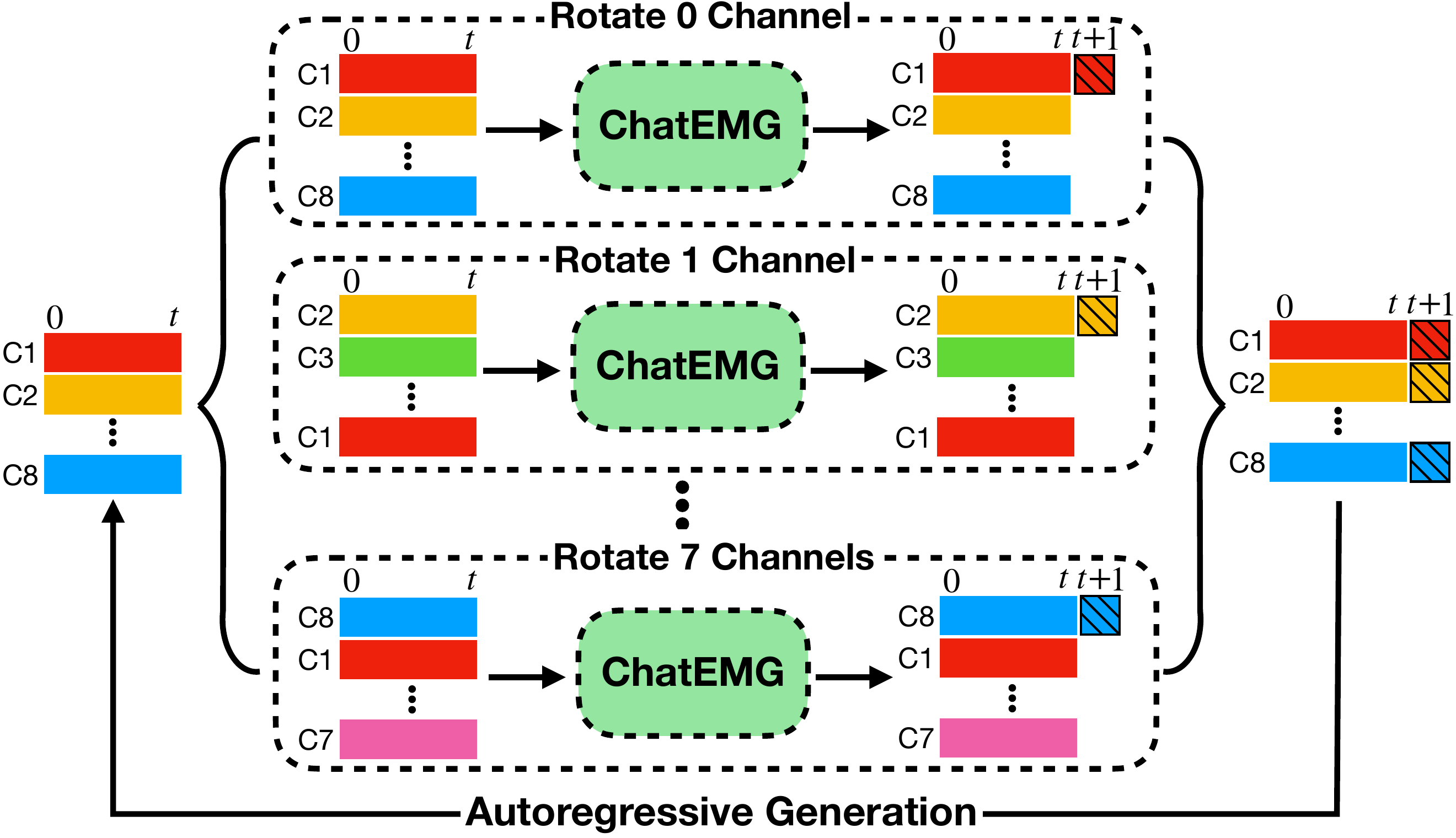}
    \vspace{-0.25in}
    \caption{\textbf{Autoregressive synthetic data generation.} ChatEMG only predicts the next EMG value for the first channel (C1) and in order to generate the complete 8-channel EMG signal, we rotate the input signals 7 times such that all other channels will become the first channel once. 
    }
    \label{fig:autoregressive}
    \vspace{-0.2in}
\end{figure}

During training, the whole output vector is compared with the ground truth next values to compute the loss. During synthetic data generation, we only use the last prediction of this output vector. In order to generate the next complete 8-channel EMG signal, we rotate the input EMG signals 7 times (one channel per time) so that each of the other 7 channels can become the first channel of the input EMG signals. We then append this newly generated signal to the input signals and continue the generation process, as shown in Fig.~\ref{fig:autoregressive}. The autoregressive nature of this architecture allows us to generate an output sequence of arbitrary length. 

\remindtext{17-6}{One key consideration for designing the ChatEMG architecture is its ability to generate diverse output given the same input signals. As a result, the output of the model is sampled from a probability distribution over a discrete vocabulary, and during training, we compute the cross-entropy loss between the predicted distribution and the one-hot label. Furthermore, we can generate an arbitrary number of ``likely'' next signals simply by repeating the sampling process. Each sequence can then be continued in an autoregressive fashion. This means that one generative model can use a single prompt to generate an arbitrary number of likely next sequences, each of arbitrary length. This is an important feature of our model, given that we will need large amounts of synthetic data in order to train downstream classifiers\footnote{As an alternative, we also tried a variational regression setup that predicts a mean and variance for the next signal, and used the reparameterization trick to backpropagate the loss, which also allows sampling an arbitrary number of completions for a single prompt. However, we found empirically that the balance between the KL (Kullback-Leibler) loss and the reconstruction loss is hard to tune, and its generated signals are not as good as the classification version.}.}


However, this architecture requires a discrete vocabulary of finite size, which is challenging given that our raw data consists of an 8-channel EMG signal. \modifiedtext{16-6}{For preprocessing, we smooth out the data using a median filter of size 9 and then bin and clip each channel of the EMG signal to be an integer between 0 and 1000.} Under this range, if we model the whole 8-channel signal as one ``token'' in our vocabulary, then our vocabulary size becomes $1000^8$, which is too large to predict a probability distribution and sample from. As a result, ChatEMG always predicts the next EMG value for only the first channel, making the vocabulary size 1000.



\subsection{Modelling Inter-channel Relationship}

Although our model predicts one channel at a time, it still considers the interchannel relationship. Our EMG armband has 8 electrodes surrounding the forearm, and each electrode covers a particular area of the forearm muscle. The interchannel relationship across different electrodes can be useful information to leverage for understanding user intent. 

The model of ChatEMG has two branches: the self branch, which takes in the first channel for which we are predicting the next EMG value, and the context branch, which takes in all 8-channel EMG signals (shown in Fig.~\ref{fig:architecture}). We use an embedding size of $N_{embed} = 256$.
The self branch uses both token and position embedding layers to compute the embedding. The context embedding block consists of 8 separate token embedding layers for each channel and one shared positional embedding layer. The channel-specific token embeddings are then summed with the shared positional embedding to create the final embedding of the context branch. Each branch has 12 attention blocks, and each block uses an 8-head attention mechanism. The output from both branches is concatenated to pass through another 3-layer fully connected network (FC).

Our model is trained with EMG sequences of length $t = 256$ (2.56 seconds at 100Hz). The dimension of the model output is $256 \times 1$, and they are the 256 predicted EMG values for the next time step of the first channel.
\remindtext{16-6}{During training, we also augment the input signals by rotating the channels seven times to simulate the rotation of the electrodes. This data augmentation strategy enables ChatEMG to be invariant with channel rotation.} 
During data generation, we sample EMG prompts of length 150 (corresponding to 1.5s) from the very limited dataset of the new condition, session, or subject and use ChatEMG to autoregressively complete the rest of the signal to a length of 256, which is the time-series length that our classification algorithms take.
\section{Experimental Setup}
The ultimate goal of orthosis is to provide meaningful functional assistance for stroke subjects, enabled by intent inferral. Thus, we examine whether the generated synthetic samples by ChatEMG can improve intent inferral accuracy. 

\subsection{Subjects}

We performed experiments with 5 chronic stroke survivors having hemiparesis and moderate muscle tone: Modified Ashworth Scale (MAS) scores $\leq$ 2 in the upper extremity. Our MAS criteria exclude subjects whose fingers are difficult to move passively --- fingers with more severe spasticity cannot be quickly extended with external forces without increasing muscle tone and risking damage to the joints. Our participants can fully close their hands but are unable to completely extend their fingers without assistance. The passive range of motion in the fingers is within functional limits. Testing was approved by the Columbia University Institutional Review Board (IRB-AAAS8104) and was performed in a clinical setting under the supervision of an occupational therapist. 


Our subjects have different hand impairments, and their Fugl-Meyer scores for upper extremity (FM-UE) vary. Subjects S1, S2, and S3 have no active finger extension (lower functioning) and have a corresponding low FM-UE score (27, 26, 26, respectively), whereas S4 and S5 have some residual active finger extension ability (higher-functioning) with a higher FM-UE score (50, 47, respectively).


\subsection{Data Collection Protocol}
\label{sec:protocol}

For each stroke subject, we collect data from two sessions on two different days using the following protocol. A \textit{session} is defined to be a single use of the device between donning and doffing the device. We intentionally keep the two sessions for each subject at least one week apart to better study the variation of the EMG signals across different days. For each session, we collect data under four different \textit{conditions}: 1) with the arm resting on a table and the orthosis motor off \{\textit{arm on table, motor off}\}, 2) with the arm resting on a table and the orthosis motor on, providing active grasp assistance \{\textit{arm on table, motor on}\}, 3) with the arm raised above the table and the orthosis motor off \{\textit{arm off table, motor off}\}, and 4) with the arm raised above the table and the orthosis motor on \{\textit{arm off table, motor on}\}.

We collect two continuous, uninterrupted \textit{recordings} for each condition, and for each recording, we instruct the subjects to open and close their hands three times by giving verbal cues of open, close, and relax. We simultaneously record the EMG signals and verbal cues as ground truth intent labels. Each verbal cue lasts for 5 seconds, and there is a relax cue between each open and close cue. For conditions where the motor is on, we move the motor approximately one second after the verbal cue is given using a dedicated button. We define each opening and closing hand completion as one round of \textit{open-relax-close motion}. Each recording then contains three open-relax-close motions. We note that this protocol is at the maximum capacity that stroke subjects can follow during a 90-minute session, and we can observe increased spasticity and fatigue at the end. 

\subsection{Assessment Scenarios}

We create different assessment scenarios (listed below) that simulate different use cases of ChatEMG, by selecting different \textit{training recordings} (recordings used to train ChatEMG) and \textit{intent inferral recordings} (recordings used to perform intent inferral evaluation). These scenarios evaluate how well ChatEMG generalizes and adapts to new conditions/sessions/subjects not seen in its training recordings. 

\subsubsection{Condition Adaptation} This scenario studies whether ChatEMG can generalize to a new condition. The training recordings are of condition \{\textit{arm on table, motor off}\} from all five subjects (including both sessions), and the intent inferral recordings are of condition \{\textit{arm off table, motor off}\}. We note that \{\textit{arm on table, motor off}\} is the most effortless condition for us to collect data in, while \{\textit{arm off table, motor off}\} is the closest condition to an ongoing functional pick-and-place task. Thus, this simulates a scenario where ChatEMG is trained on data collected in the effortless condition and used to generate synthetic samples for a drastically different but realistic condition. 

\subsubsection{Session Adaptation} This scenario pertains to the signal variation across different use sessions, and it seeks to simulate using the orthosis on a subject seen previously in a different session on a different day. ChatEMG is trained on recordings of the first session from all subjects, and the intent inferral recordings are those of the second session.

\subsubsection{Subject Adaptation} This scenario simulates onboarding new subjects. We conduct five separate experiments, each one simulating the onboarding of one holdout subject, given that we have seen the other four. In each experiment, we train ChatEMG using all the recordings from the other subjects (including both sessions), and the intent inferral recordings are those of the holdout subject. When adapting to a new subject, it is also implicitly adapting to a new session. However, in our session adaption experiments, we assume it is a different session of a previously seen subject. 

\addedtext{17-4}{We use a subset of the training recordings (around 314K samples of size 256 by 8) to train ChatEMG, and we use the remaining training recordings (around 204K samples) as the validation set. We early stop the generative training before the validation loss increases to avoid overfitting.}

\vspace{-0.1in}
\subsection{Intent Inferral Classifiers}


ChatEMG is classifier-agnostic, and the generated synthetic samples can be integrated with the training set of any classifiers. We study three types of classifiers: linear discriminant analysis (LDA), random forests (RF), and Transformer. They are popular in the biomedical literature and cover both classic machine-learning algorithms and high-capacity neural networks. We feed into each classifier a time series of length 256 (2.56s), in the shape of 256 by 8. The EMG signals are flattened into a single vector for LDA and RF. We use a single 4-head attention block followed by a 3-layer multilayer perception (MLP) for the Transformer classifier. \addedtext{16-9}{We perform the same preprocessing techniques as training the ChatEMG model, and as an additional step, we normalize the EMG signals into the range of $[-1, 1]$.}

\subsection{Baselines}


The intent inferral evaluation is done on individual intent inferral recordings. For each recording, we assume only a small \textit{support set} (i.e., the first open-relax-close motion of the recording) is available for training the classifier. The support set simulates the limited new training samples from a new condition, session, or subject, denoted by $\mathcal{D}_{\text{new}}^{\text{synth}}$ in Sec.~\ref{sec:overview}. We then test the classifier's accuracy using the \textit{query set} (i.e., the second and third open-relax-close motions). 

\subsubsection{Self} This method trains the intent inferral classifier using only the support set of the intent inferral recordings. 
\subsubsection{Fine-tune} This method pre-trains the classifier using all the training recordings of ChatEMG and then fine-tunes on the support set of the intent inferral recordings. This baseline ensures the training data for ChatEMG is also accessible for a fair comparison.
\subsubsection{ChatEMG} This is our proposed method. We repetitively sample prompts of size 150 (1.5s) from the small support set and leverage ChatEMG models to expand the prompts to size 256 (2.56s). These synthetic samples are then combined with the original support set to train intent classifiers. For each intent, we add 1000 synthetic samples.

\section{Results and Discussion}

In this section, we first discuss the intent inferral accuracy and then analyze the synthetic samples. Finally, we show that ChatEMG can help improve the performance of functional pick-and-place tasks in real-world hospital testing. 
Visit our project website at \url{https://jxu.ai/chatemg} for hospital testing demonstrations and additional information.

\subsection{Intent Inferral Performance}

\begin{table*}
    \caption{\textbf{Condition adaptation experiment results.}}
    \renewcommand{\arraystretch}{0.6}
    \centering
    \begin{tabular}{c|c|cccccc|c}
    \toprule
    & & \textbf{S1} & \textbf{S2} & \textbf{S3} & \textbf{S4} & \textbf{S5} & \textbf{Average} & \begin{tabular}[c]{@{}c@{}}\textbf{$p$-value w\slash}\\\textbf{\textit{ChatEMG}} \end{tabular} \\
    \midrule
    \multirow{3}{*}{LDA} & \textit{Self} & $0.37 \pm 0.15$ & $0.54 \pm 0.07$ & $0.46 \pm 0.10$ & $0.64 \pm 0.22$ & $0.36 \pm 0.07$ & $0.48$ & $\mathbf{3\mathrm{\textbf{e}}{-4}}$ \\
    & \textit{Fine-tune} & $0.34 \pm 0.12$ & $0.64 \pm 0.10 $ & $0.54 \pm 0.05$ & $0.71 \pm 0.05$ & $0.48 \pm 0.15$ & $0.54$ & $\mathbf{4\mathrm{\textbf{e}}{-4}}$\\
    & \textit{ChatEMG} & $0.45 \pm 0.06$ & $0.69 \pm 0.02$ & $0.70 \pm 0.09$ & $0.90 \pm 0.04$ & $0.68 \pm 0.04$ & $\textbf{0.68}$ & --- \\
    \midrule
    \multirow{3}{*}{RF} & \textit{Self} & $0.52 \pm 0.07$ & $0.72 \pm 0.08$ & $0.64 \pm 0.10$ & $0.90 \pm 0.08$ & $0.77 \pm 0.06$ & $0.71$ & $1\mathrm{e}{-1}$\\
    & \textit{Fine-tune} & $0.53 \pm 0.11$ & $0.68 \pm 0.06$ & $0.63 \pm 0.04$ & $0.89 \pm 0.08$ & $0.64 \pm 0.17$ & $0.67$ & $\mathbf{5\mathrm{\textbf{e}}{-2}}$\\
    & \textit{ChatEMG} & $0.53 \pm 0.14$ & $0.70 \pm 0.11$ & $0.71 \pm 0.11$ & $0.92 \pm 0.04$ & $0.77 \pm 0.05$ & $\textbf{0.73}$ & --- \\
    \midrule
    \multirow{3}{*}{Transformer} & \textit{Self} & $0.55 \pm 0.05$ & $0.64 \pm 0.05$ & $0.58 \pm 0.10$ & $0.84 \pm 0.06$ & $0.73 \pm 0.07$ & $0.67$ & $\mathbf{3\mathrm{\textbf{e}}{-2}}$ \\
    & \textit{Fine-tune} & $0.57 \pm 0.06$ & $0.64 \pm 0.05$ & $0.70 \pm 0.03$ & $0.88 \pm 0.04$ & $0.68 \pm 0.06$ & $0.69$ & $2\mathrm{e}{-1}$\\
    & \textit{ChatEMG} & $0.60 \pm 0.15$ & $0.72 \pm 0.08$ & $0.65 \pm 0.01$ & $0.86 \pm 0.04$ & $0.78 \pm 0.04$ & $\textbf{0.72}$ & --- \\
    \bottomrule
    \end{tabular}
    \label{tab:cross_condition}
    \vspace{-0.1in}
\end{table*}

\begin{table*}
    \caption{\textbf{Session adaptation experiment results.}}
    \footnotesize
    \renewcommand{\arraystretch}{0.6}
    \centering
    \begin{tabular}{c|c|cccccc|c}
    \toprule
    & & \textbf{S1} & \textbf{S2} & \textbf{S3} & \textbf{S4} & \textbf{S5} & \textbf{Average} & \begin{tabular}[c]{@{}c@{}}\textbf{$p$-value w\slash}\\\textbf{\textit{ChatEMG}} \end{tabular} \\
    \midrule
    \multirow{3}{*}{LDA} & \textit{Self} & $0.40 \pm 0.16$ & $0.50 \pm 0.09$ & $0.52 \pm 0.07$ & $0.50 \pm 0.13$ & $0.81 \pm 0.05$ & $0.54$ & $\mathbf{6\mathrm{\textbf{e}}{-5}}$ \\
    & \textit{Fine-tune} & $0.64 \pm 0.04$ & $0.49 \pm 0.07$ & $0.65 \pm 0.05$ & $0.65 \pm 0.02$ & $0.64 \pm 0.13$ & $0.61$ & $\mathbf{1\mathrm{\textbf{e}}{-2}}$\\
    & \textit{ChatEMG} & $0.54 \pm 0.07$ & $0.55 \pm 0.05$ & $0.64 \pm 0.07$ & $0.77 \pm 0.08$ & $0.83 \pm 0.08$ & $\textbf{0.67}$ & --- \\
    \midrule
    \multirow{3}{*}{RF} & \textit{Self} & $0.55 \pm 0.08$ & $0.57 \pm 0.07$ & $0.77 \pm 0.11$ & $0.82 \pm 0.13$ & $0.77 \pm 0.06$ & $0.69$ & $1\mathrm{e}{-1}$ \\
    & \textit{Fine-tune} & $0.58 \pm 0.06$ & $0.62 \pm 0.02$ & $0.74 \pm 0.16$ & $0.72 \pm 0.16$ & $0.78 \pm 0.05$ & $0.69$ & $2\mathrm{e}{-1}$ \\
    & \textit{ChatEMG} & $0.57 \pm 0.07$ & $0.66 \pm 0.03$ & $0.72 \pm 0.15$ & $0.82 \pm 0.13$ & $0.79 \pm 0.07$ & $\textbf{0.71}$ & --- \\
    \midrule
    \multirow{3}{*}{Transformer} & \textit{Self} & $0.66 \pm 0.06$ & $0.53 \pm 0.05$ & $0.77 \pm 0.09$ & $0.67 \pm 0.15$ & $0.72 \pm 0.15$ & $0.67$ & $\mathbf{2\mathrm{\textbf{e}}{-2}}$ \\
    & \textit{Fine-tune} & $0.72 \pm 0.06$ & $0.57 \pm 0.12$ & $0.78 \pm 0.12$ & $0.79 \pm 0.09$ & $0.79 \pm 0.08$ & $\textbf{0.73}$ & $5\mathrm{e}{-1}$ \\
    & \textit{ChatEMG} & $0.64 \pm 0.12$ & $0.67 \pm 0.05$ & $0.77 \pm 0.14$ & $0.79 \pm 0.04$ & $0.77 \pm 0.12$ & $\textbf{0.73}$ & --- \\
    \bottomrule
    \end{tabular}
    \label{tab:cross_session}
    \vspace{-0.1in}
\end{table*}

\begin{table*}
    \caption{\textbf{Subject adaptation experiment results.}}
    \footnotesize
    \renewcommand{\arraystretch}{0.6}
    \centering
    \begin{tabular}{c|c|cccccc|c}
    \toprule
    & & \textbf{S1} & \textbf{S2} & \textbf{S3} & \textbf{S4} & \textbf{S5} & \textbf{Average} & \begin{tabular}[c]{@{}c@{}}\textbf{$p$-value w\slash}\\\textbf{\textit{ChatEMG}} \end{tabular} \\
    \midrule
    \multirow{3}{*}{LDA} & \textit{Self} & $0.33 \pm 0.06$ & $0.61 \pm 0.11$ & $0.51 \pm 0.11$ & $0.52 \pm 0.26$ & $0.65 \pm 0.18$ & $0.51$ & $\mathbf{1\mathrm{\textbf{e}}{-2}}$ \\
    & \textit{Fine-tune} & $0.60 \pm 0.08$ & $0.62 \pm 0.10$ & $0.49 \pm 0.06$ & $0.74 \pm 0.03$ & $0.40 \pm 0.17$ & $0.56$ & $3\mathrm{e}{-1}$ \\
    & \textit{ChatEMG} & $0.37 \pm 0.18$ & $0.63 \pm 0.11$ & $0.56 \pm 0.16$ & $0.68 \pm 0.15$ & $0.70 \pm 0.14$ & $\textbf{0.58}$ & --- \\
    \midrule
    \multirow{3}{*}{RF} & \textit{Self} & $0.60 \pm 0.01$ & $0.64 \pm 0.02$ & $0.59 \pm 0.02$ & $0.80 \pm 0.08$ & $0.71 \pm 0.07$ & $0.66$ & $3\mathrm{e}{-1}$ \\
    & \textit{Fine-tune} & $0.54 \pm 0.02$ & $0.65 \pm 0.02$ & $0.55 \pm 0.12$ & $0.78 \pm 0.07$ & $0.75 \pm 0.13$ & $0.65$ & $\mathbf{5\mathrm{\textbf{e}}{-2}}$ \\
    & \textit{ChatEMG} & $0.52 \pm 0.01$ & $0.65 \pm 0.06$ & $0.55 \pm 0.12$ & $0.86 \pm 0.06$ & $0.80 \pm 0.08$ & $\textbf{0.67}$ & --- \\
    \midrule
    \multirow{3}{*}{Transformer} & \textit{Self} & $0.57 \pm 0.05$ & $0.63 \pm 0.03$ & $0.63 \pm 0.06$ & $0.72 \pm 0.04$ & $0.65 \pm 0.16$ & $0.64$ & $\mathbf{1\mathrm{\textbf{e}}{-2}}$ \\
    & \textit{Fine-tune} & $0.56 \pm 0.08$ & $0.64 \pm 0.02$ & $0.60 \pm 0.10$ & $0.68 \pm 0.08$ & $0.69 \pm 0.04$ & $0.64$ & $\mathbf{4\mathrm{\textbf{e}}{-2}}$ \\
    & \textit{ChatEMG} & $0.58 \pm 0.13$ & $0.64 \pm 0.07$ & $0.67 \pm 0.06$ & $0.76 \pm 0.03$ & $0.74 \pm 0.14$ & $\textbf{0.68}$ & --- \\
    \bottomrule
    \end{tabular}
    \label{tab:cross_subject}
    \vspace{-0.1in}
\end{table*}

The results are shown in Table~\ref{tab:cross_condition},~\ref{tab:cross_session} and~\ref{tab:cross_subject}.
For each subject, we evaluate three intent inferral recordings, and we present the average accuracy and one standard deviation. ChatEMG is able to improve the average intent inferral accuracy across five subjects for all classifiers under all assessment scenarios. This shows that ChatEMG can successfully generalize to a new condition, session, or subject despite not seeing them in its training recordings. 

\addedtext{16-11}{Despite that the improvement in intent inferral accuracy is consistent across 17/18 comparisons between ChatEMG and Fine-tune/Self, we further investigate the statistical significance of such improvement by performing a one-sided Wilcoxon rank-sum test on the results aggregated across all subjects (three intent inferral recordings per subject). We choose a non-parametric statistical test because we do not assume an underlying normal distribution. We report the computed $p$-values for pairwise differences between our ChatEMG and the other methods. With a commonly used hypothesis threshold of $\alpha = 5\mathrm{e}{-2}$, 11/17 improvements (\mbox{$p$-values} in bold) are statistically significant.} 

We notice that if trained only on the small support set (Self), RF tends to have the highest performance, while when the classifier has access to larger datasets (Fine-tune or ChatEMG), Transformer tends to perform better. This matches our intuition that larger-capacity models can realize their potential only when given enough data, and ChatEMG achieves that through synthetic data generation. 

Subject adaptation is the most difficult scenario with the lowest intent inferral accuracy. It simulates the scenarios of onboarding a new stroke subject, and we only collect one round of open-relax-close motion from this new subject as our support set. It is the most tricky scenario because variation in EMG signals is larger among different subjects than among different conditions or sessions of the same subject. However, ChatEMG can still understand the broad signal patterns of different intents from past subjects and apply that knowledge by generating synthetic samples conditioned on prompts from the new subject. S4 and S5 tend to have higher intent inferral accuracy than S1, S2, and S3, which matches the hand-functionality measured by the FM-UE scores.

\subsection{Synthetic Sample Visualization and Analysis}

\begin{figure}
    \centering
    \includegraphics[width=\linewidth]{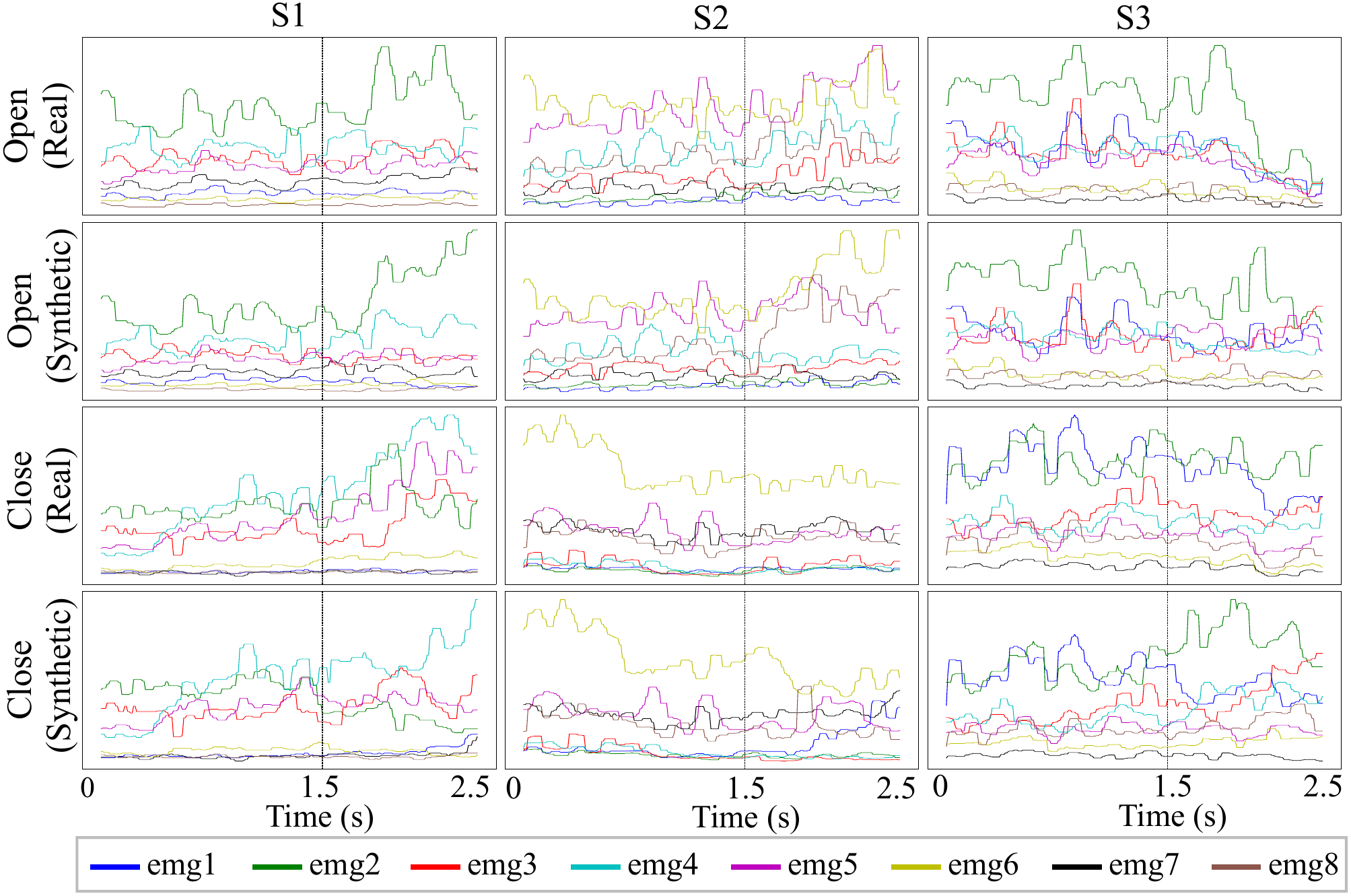}
    \vspace{-0.2in}
    \caption{\textbf{Comparison between the real and synthetic samples on open and close intents of subjects S1, S2, and S3.} The vertical line indicates the switch from the provided prompt to the generated synthetic sequence. These samples also demonstrate the significant variations in EMG signals across different stroke subjects.}
    \label{fig:samples}
    \vspace{-0.2in}
\end{figure}

\begin{figure}
    \centering
    \includegraphics[width=\linewidth]{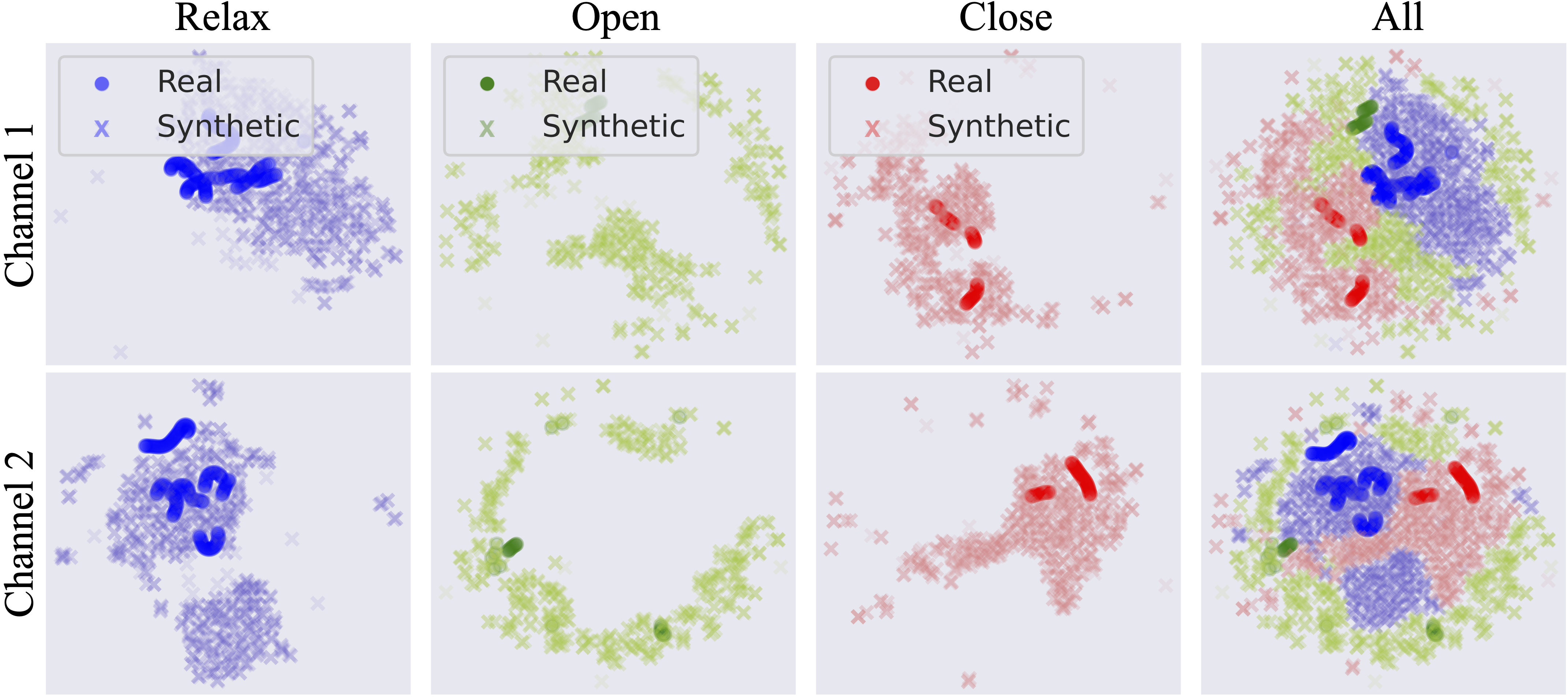}
    \vspace{-0.2in}
    \caption{\textbf{t-SNE visualization.} We compare the t-SNE embedding space of the first 2 EMG channels between the synthetic samples and real samples from the same recording of subject S4.}
    \label{fig:tsne}
    \vspace{-0.2in}
\end{figure}


We show generated samples for lower-functioning subjects S1, S2, and S3 in Fig.~\ref{fig:samples}. The first 150 steps (1.5s) are the sampled prompts from the limited dataset and are identical between real and synthetic data. For the synthetic samples, ChatEMG generates the last 106 steps (1.06s). When presented with both plots in parallel, without knowing in advance, it is very challenging to identify the synthetic one. There is no significant transition at 1.5s when the sequence switches from real to synthetic. This shows that ChatEMG can capture the characteristics of EMG signals, such as the amplitude, frequency, fluctuation pattern, etc. ChatEMG can also maintain the relative position of different channels very well. \addedtext{6-3}{We compute the normalized root mean squared error (NRMSE) between the synthetic and real samples across all subjects and intents. The NRMSE for close, open, and relax intents across all subjects are 6\%, 5\%, and 3\%, respectively.} 

More importantly, ChatEMG not only learns to babble EMG signals by following the previous trends,
but it also learns to reproduce common trends that do not show up in the prompt at all. For example, in the prompt of S1's open sample, there is no indication of the green channel (emg2) going up, but the generated sequence shows such a trend, which turns out to be correct. 

We further visualize the generated samples of subject S4 in a low-dimensional space using \mbox{t-SNE~\cite{van2008visualizing}}, shown in Fig.~\ref{fig:tsne}. For each class, we generate 1000 synthetic samples using the support set and then randomly select 100 samples from the query set of the same recording. We embed each channel of the 256-step EMG sequence separately into a 2D space, and visualize that of the first two channels. We observe that: (1) The synthetic samples of different intents are very separable from each other, meaning that ChatEMG captures distinct patterns of different intents. (2) The embedding space of the generated samples almost always covers the real samples from the query set. This shows that ChatEMG captures the distribution of the test samples correctly. Thus, adding these synthetic samples to the limited support set can improve the intent inferral accuracy on the query set.

\subsection{Integration in Complete Subject Protocol}

We deploy ChatEMG to help an unseen stroke subject complete a functional pick-and-place task using a robotic hand orthosis, as shown in Fig.~\ref{fig:teaser}. 
We integrate the pipeline of collecting a limited support set, using ChatEMG to generate synthetic samples, and training Tranformer classifiers within a single hospital session.
Visit our project website for video demonstrations.
This preliminary experiment uses the ChatEMG models trained with data from S1, S2, S4, and S5, and excludes data from the test subject S3. Without adding the synthetic data, classifiers trained with only one open-close-relax motion cannot predict the open intent at all. However, when the classifier is trained with synthetic sample augmentation, S3 can complete multiple rounds of pick-and-place tasks. These qualitative results suggest that the improvement in classification accuracy can translate to the improvement of meaningful daily functional tasks.
\section{Conclusion}

We propose ChatEMG, an autoregressive generative model that can generate synthetic EMG signals conditioned on prompts. 
ChatEMG allows us to collect only a very small dataset from the new condition/session/subject and expand it with synthetic samples. ChatEMG learns the broad behavior of forearm EMG signals from a vast corpus of previous data while remaining context-specific via prompting. We show that these synthetic samples are classifier-agnostic and can improve the intent inferral accuracy of different types of classifiers. We are the first to deploy an intent classifier trained partially on synthetic data on a hand orthosis to help an unseen stroke subject complete pick-and-place tasks, showing that the improvement in classification accuracy can lead to improvement in meaningful functional tasks.

\bibliographystyle{IEEEtran}
\bibliography{references}

\end{document}